%% file: main.tex
\documentclass[10pt,twocolumn,letterpaper]{article}

 \usepackage{cvpr}              % To produce the CAMERA-READY version
\input{preamble}

\definecolor{cvprblue}{rgb}{0.21,0.49,0.74}
\usepackage[pagebackref,breaklinks,colorlinks,citecolor=cvprblue]{hyperref}
\usepackage[rightcaption]{sidecap}
\usepackage[accsupp]{axessibility} % Improves PDF readability for those with visual impairments.
\newtheorem{definition}{Definition}[section] 
\def\confName{CVPR}
\def\confYear{2024}

\title{Improving Single Domain-Generalized Object Detection: A Focus on Diversification and Alignment}
\author{Muhammad Sohail Danish\textsuperscript{1}
\and
Muhammad Haris Khan\textsuperscript{1}
\and
Muhammad Akhtar Munir\textsuperscript{1, 2} 
\and 
M. Saquib Sarfraz\textsuperscript{3, 4} \qquad\qquad
Mohsen Ali\textsuperscript{2}
\and
\small{
\textsuperscript{1}Mohamed bin Zayed University of Artificial Intelligence,   \textsuperscript{2}Information Technology University of Punjab, } \\
\small{\textsuperscript{3}Mercedes-Benz Tech Innovation, \textsuperscript{4}Karlsruhe Institute of Technology}\\
{\tt\small \href{muhammad.sohail@mbzuai.ac.ae}{muhammad.sohail@mbzuai.ac.ae}, \href{muhammad.haris@mbzuai.ac.ae}{muhammad.haris@mbzuai.ac.ae} }
}

\begin{document}
\maketitle
\input{sec/0_abstract}    
\input{sec/1_intro}
\input{sec/2_relatedwork}

\input{sec/3_method}

\input{sec/4_experiments}

\input{sec/5_conslusion}
\input{sec/X_suppl}

\clearpage
{
    \small
    \bibliographystyle{ieeenat_fullname}
    \bibliography{egbib}
}


\end{document}

%% file: preamble.tex
\usepackage[table,xcdraw]{xcolor}

%% file: sec/0_abstract.tex
\begin{abstract}

In this work, we tackle the problem of domain generalization for object detection, specifically focusing on the scenario where only a single source domain is available. We propose an effective approach that involves two key steps: diversifying the source domain and aligning detections based on class prediction confidence and localization.
Firstly, we demonstrate that by carefully selecting a set of augmentations, a base detector can outperform existing methods for single domain generalization by a good margin. This highlights the importance of domain diversification in improving the performance of object detectors.
Secondly, we introduce a method to align detections from multiple views, considering both classification and localization outputs. This alignment procedure leads to better generalized and well-calibrated object detector models, which are crucial for accurate decision-making in safety-critical applications.
Our approach is detector-agnostic and can be seamlessly applied to both single-stage and two-stage detectors.
To validate the effectiveness of our proposed methods, we conduct extensive experiments and ablations on challenging domain-shift scenarios. The results consistently demonstrate the superiority of our approach compared to existing methods.
Our code and models are available at:  \href{https://github.com/msohaildanish/DivAlign}{https://github.com/msohaildanish/DivAlign}.

\end{abstract}

%% file: sec/1_intro.tex
\section{Introduction}
\label{sec:intro}

In recent years, we have witnessed remarkable performance improvements in supervised object detection \cite{ren2015faster,redmon2016you, liu2016ssd, tian2019fcos}. The success of these methods rely on the assumption that the training and testing data are sampled from the same distribution. However, in many real-world applications, such as autonomous driving, this assumption is often violated and these object detectors usually show degraded performance due to a phenomenon called domain-shift \cite{chen2018domain,saito2019strong,chen2020harmonizing}. Shifts in real-world domains are typically caused by environmental alterations, like varying weather and time conditions. These changes manifest in diverse contrasts, brightness levels, and textures among others.

A prominent line of research that attempts to alleviate the impact of domain-shift is known as unsupervised domain adaptation (UDA) \cite{inoue2018cross, zhu2019adapting, saito2019strong,hsu2020every, zheng2020cross, xu2020cross, munir2021ssal}. Given the labeled data from the source domain and unlabelled data from the target domain, the aim of UDA methods is to align the source and target data distributions so that the trained model can generalize well to the target domain \cite{zhao2020review}. An obvious limitation of UDA methods is that they require pre-collecting data and re-training the model for different target domains. 
Collecting data, even without annotation, for all possible domain shifts and training the model when shift happens is difficult and sometime not possible. 

To cope with the domain-shift problem, a more realistic albeit challenging problem is domain generalization \cite{muandet2013domain, li2017deeper, li2018domain, carlucci2019domain, li2019episodic, dou2019domain, wang2020learning, huang2020self, khan2021mode}. The goal of \textit{domain generalization} (DG) is to learn a generalizable model typically from multiple source domains, available during training, that can perform well on the unseen target domain. A dominant class of DG methods attempts to learn domain-invariant feature space across multiple source domains \cite{balaji2018metareg, li2018learning, li2018domain, dou2019domain}. The performance of these methods is sensitive to the number of available source domains \cite{choi2021robustnet, zhou2022domain}.  In many realistic scenarios, acquiring labeled data from multiple source domains is often costly and time-consuming, eventually restricting the potential utilization of such methods. For these reasons, generalizing from a single-source domain is a more practical setting.

Recent surveys \cite{zhou2022domain, koh2021wilds} reveal that very little work has been done on DG for object detection \cite{liu2020towards, wu2022single}, despite the fact that object detectors occupy an important position in many safety-critical and security-sensitive applications e.g., autonomous driving. An object detector should be able to provide accurate as well as calibrated detections in different out-domain scenarios. Towards this end, we study the problem of single-domain generalized object detection (Single-DGOD). In the Single-DGOD setting, a single source domain is available for training, and the goal is to learn an object detector that can generalize well to multiple unseen target domains. Note that, many existing DG methods cannot be applied to solve this task since both multiple source domains and domain-level annotations are unavailable.

We are inspired by the class of DG methods for classification \cite{qiao2020learning, volpi2019addressing, zhao2020maximum} that show that simulating novel domains during model training allows segregating domain-specific and semantics-oriented features. Consequently, the spurious correlations between the input and model predictions are suppressed. We begin by leveraging this idea of augmenting our input examples to realize the simulation of novel domains, thereby increasing the diversity of single-source training domain. In particular, we use the common off-the-shelf visual corruptions coupled with a simple training scheme to build a strong Single-DGOD baseline. The augmentation strategy aims to disrupt surface-level statistical patterns that are specific to a particular domain while preserving high-level semantic concepts that are common across domains.
In addition to diversifying the single source domain, we propose a novel approach that aligns detection across different  views of the image. The introduced method aligns both the class prediction distributions and predicted bounding box coordinates between augmented views of the image. By ensuring such a consistency of detections, we show that this not only helps in better out-domain detections but importantly also results in better calibrated models.

To summarize, our work makes two main contributions:
\begin{itemize}
\item We show that, using a set of carefully selected augmentations, the base model can surpass the performanace of state-of-the-art published single-DGOD methods. To our knowledge, achieving such a strong baseline with known augmentation methods has not been shown before in the single domain object detection generalization.  
\item Our second contribution is an effective classification and localization alignment method that is shown to further improve detections in unseen domain and results in implicitly well calibrated detector model.

\end{itemize}
Our approach is detector-agnostic and can be seamlessly applied to both single-stage and two-stage object detectors. 
\begin{figure*}[t]
\centering
\includegraphics[width=0.95\linewidth]{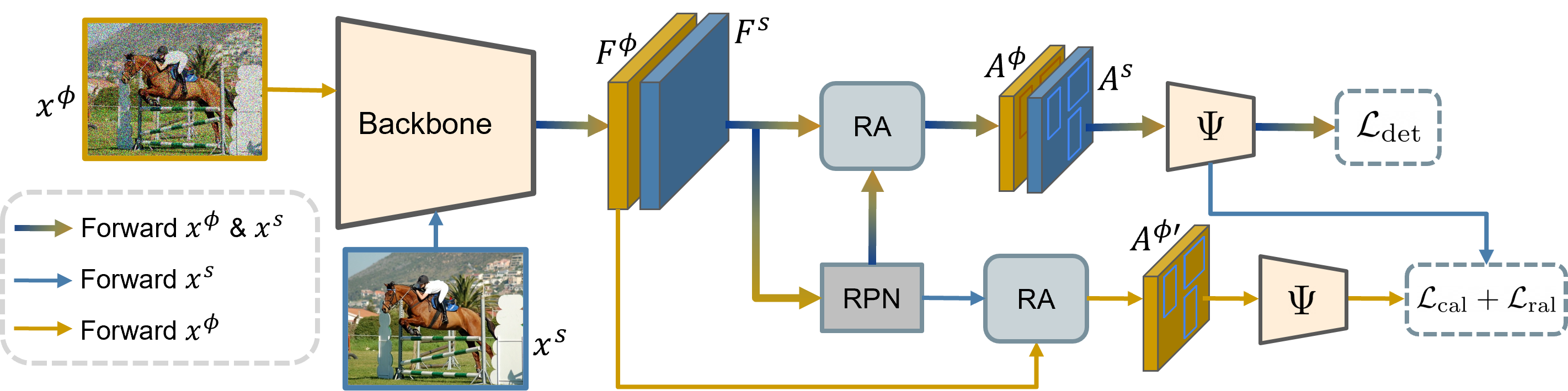}
\caption{Overall architecture of our proposed method. At the core is a baseline detector, Here a two-stage detector Faster-RCNN\cite{ren2015faster} is depicted, comprising of backbone, region proposal network (RPN), and ROI alignment (RA). To improve the single domain generalization of the baseline detector, we propose to diversify the single source domain and also align the diversified views by minimizing losses at both classification and regression outputs.}\label{fig:architecture_new1}
\end{figure*}

%% file: sec/2_relatedwork.tex
\section{Related Work}\label{sec:Related Work}

\noindent\textbf{Domain adaptive object detection:}
Several domain adaptive detection methods aim to reduce the pixel-level or feature-level gap between the source and target domain \cite{li2020deep,oza2021unsupervised}. For instance, the works of 
\cite{chen2020harmonizing,deng2021unbiased,hsu2020progressive,inoue2018cross,zhang2021rpn, Sultani_2022_CVPR} utilized the labeled images generated from CycleGAN \cite{zhu2017unpaired} to achieve pixel-level consistency. Likewise, \cite{chen2020harmonizing,chen2018domain,deng2021unbiased, he2019multi,saito2019strong, shen2019scl} proposed various mechanisms to progressively obtain a feature-level consistency. A common component of these methods is domain adversarial network \cite{ganin2016domain} for achieving feature alignment. The work of \cite{chen2018domain} performed domain alignment on both global features (backbone features) and local features (Instance-level RoI features). Since then, several methods aimed to further improve these alignments by leveraging multi-scale \cite{he2019multi}, contextual \cite{chen2020harmonizing,saito2019strong}, spatial attention \cite{li2020spatial}, category attention \cite{vs2021mega}, and cross-domain topological relations \cite{chen2021dual} information. A few methods resorted to pseudo-labelling of instances to include class-level discriminative information. For instance, \cite{inoue2018cross} constructed pseudo-labels by putting a threshold on top-1 accuracy over the predicted class of a detection. Towards refining the noisy pseudo-labels, \cite{khodabandeh2019robust} developed a robust training mechanism coupled with an independent classification module. With a similar goal, \cite{kim2019self} proposed a weak self-training method and an adversarial background regularization module. 

Recently, \cite{munir2021ssal} developed an uncertainty-guided criterion for selecting pseudo-labels in the target domain. We note that almost all of the aforementioned methods require access to both source and (unlabelled) target domain data coupled with an adaptation step to perform well in the target domain. As a consequence, these methods cannot be used to solve single-domain generalized object detection. 

\noindent\textbf{Single source domain generalization:} A more challenging and recently proposed setting in DG is single source domain generalization. It is a more practical setting compared to multi-source DG since the availability of data from multiple different domains is a difficult requirement to fulfill for many real-world applications. The goal is to learn a generalized model using only a single source domain that can perform well on many unseen target domains. An intuitive approach for this setting is to increase the diversity of single source data by proposing different data augmentation techniques. 
Qiao \etal. \cite{qiao2020learning} leveraged adversarial training to construct difficult examples which facilitate model generalization. 
Earlier attempts involve training a label classifier and a domain classifier in a joint manner by their corresponding perturbations Similarly‑\cite{shankar2018generalizing} proposed to impose wasserstein constraint in the feature space to generate adversarial samples from synthetic target distribution \cite{volpi2018generalizing}. 

Recently, Wu \etal. \cite{wu2022single} proposed a cyclic-disentangled self-distillation approach for single domain-generalized object detection. 
Wang \etal. \cite{wang2021learning} proposed a style-complement module that synthesizes examples from domains complementary to the source domain. Towards exploring the impact of the normalization layer on DG performance, Wang \etal. \cite{fan2021adversarially} developed a generic normalization approach that learns both the standardization and rescaling statistics by neural networks. 

The aforementioned methods have shown promising results for the image classification task, however, they are not readily applicable to object detection since it has both classification and localization.  

\noindent\textbf{Domain diversification for domain generalization:} Several DG methods for object recognition have proposed various data augmentation techniques in an effort to improve the diversity of source domain(s). By using a CNN generator with different losses, Zhou \etal. \cite{zhou2020learning} synthesized novel examples from simulated domains. 
A few methods attempt to suppress the intrinsic style bias of CNNs. For instance, to focus on the contents and styles of images, \cite{nam2021reducing} proposed content-biased and style-biased networks, respectively. To constantly seek diverse styles, \cite{kang2022style} proposed a style synthesis process that is formulated as a monotone submodular optimization.
Among the more recent methods, \cite{fan2023towards} explores diversifying source domain by perturbing channel statistics of low-level features for single domain-generalized object detector. In~\cite{Vidit_2023_CVPR} the authors introduced the semantic augmentation of images by utilizing text-prompts relevant to potential target domain concepts in a vision-language model framework to improve the robustness of object detectors in unseen target domains. Along similar lines, \cite{fahes2023poda} adapted the source model using textual descriptions of the target domain in a zero-shot manner to enhance the cross-domain generalizability in object detectors. Note that, both \cite{Vidit_2023_CVPR} and \cite{fahes2023poda} incorporate the prior target domain knowledge using textual prompts. However, we follow domain generalization settings and leverage no prior knowledge about the target domain. Further, we diversify the source domain in input space and then align the detections in both original and diversified images.

%% file: sec/3_method.tex
\section{Proposed Methodology}\label{sec:proposed_methodology}

\subsection{Preliminaries}\label{subsection:preliminaries}

\noindent\textbf{Problem Settings:} Let $\mathcal{D}_s$ be the single source domain containing 
$N^s$ labelled training examples 
$\{(\mathbf{x}^s_i, \mathbf{y}^s_i) \}_{i=1}^{N^s}$, where $\mathbf{x}_i \in \mathbb{R}^{H \times W \times C}$ is an image and $\mathbf{y}_i = \{k_i, \mathbf{b}_i\}$ is the corresponding label with bounding box coordinates $\mathbf{b}_i \in \mathbb{R}^4$ and the associated class category $k_i \in \{ 1,...,K \}$. $K$ is the number of class categories and $H$, $W$ and $C$ represent the height, width, and number of channels, respectively. Let $\{\mathcal{D}_t\}_{t=1}^{T}$ be the set of $T$ (unseen) target domains. Our goal is to learn an object detector using training examples from $\mathcal{D}_s$ that generalizes well on test examples from $\mathcal{D}_t$. We assume that both $\mathcal{D}_s$ and $\mathcal{D}_t$ share the same label space.

\noindent\textbf{Object Detection:} Let $\mathcal{F}_{det}$ be an object detection model, that takes input image $\mathbf{x}$ and outputs $Z$ detections $\{(P(\hat{k}_n|\mathbf{x}), \hat{\mathbf{b}}_n)\}^{Z}_{n=1}$, where $\hat{k}_n$ denotes the class and bounding box $\hat{\mathbf{b}}_n$ predictions for $n^{th}$ proposal, respectively, 

We take the Faster R-CNN \cite{ren2015faster} model as $\mathcal{F}_{det}$ to detail our method, but note that our method is detector-agnostic and can be easily applied to other object detection paradigms, as shown in our results. We first overview Faster R-CNN to make the method section self-contained.   
Faster R-CNN is a popular and widely used two-stage object detector. The first stage comprises of a backbone feature extractor layer which generates a feature map and a region proposal network (RPN) which predicts a set of object proposals or regions of interest (RoIs) from the feature map.
The second stage begins with ROI-Alignment, which extracts a fixed-size feature representation for each ROI and then feeds it to a classifier and a regressor to predict class confidence scores and bounding box coordinates, respectively.

\noindent\textbf{Single domain-generalized object detection:} In this case, we assume that the object detection $\mathcal{F}_{det}$ has been trained on the examples taken from the source domain $\mathcal{D}_s$ and then tested on examples from $\{\mathcal{D}_t\}_{t=1}^{T}$. 
Akin to image classification models, object detection models like 
Faster-RCNN, which exhibits state-of-the-art performance when testing data and training data are from the same domain (i.e. in-domain scenario), but suffers from performance degradation when domains are different (i.e. out-of-domain scenario) (Figure \ref{fig:voc_map_ece}).
The presence of multiple domains during training penalizes the object detector in learning shortcuts which are typically domain-specific and instead encourages learning domain-invariant features. 

Nevertheless, the multi-domain data with the ground truth are difficult requirements to fulfill in several realistic settings and thus we are reduced to a single-domain generalization problem.

\noindent\textbf{Domain invariant object detection:}
We first re-formulate the definition of the domain invariance for a single source domain-generalized object detector by including a classifier and a regressor. 
Following \cite{Xiao2021bayesian}, let's consider a visual corruption function $\phi(.)$ as a domain transformation function, in (continuous) domain space $\mathfrak{D}$, which takes a source image from $\mathcal{D}_s$ and transforms it into a different domain $\mathcal{D}_{\phi}$ where $\phi \sim \Phi$ 
(is a set of possible transformations). 

By making this assumption, we define domain-invariance for single domain- generalized object detector as follows:
  \begin{definition}[Domain Invariance for Object Detection] \label{def:invariant_def}

Assuming that, for an input image $\mathbf{x}$, an object detection model $\mathcal{F}_{det}$ predicts class probability distribution $\hat{\mathbf{p}}_n$ and bounding box coordinates $\hat{\mathbf{b}}_n \in \mathbb{R}^4$ for the $n^{th}$ proposal.
Let $\mathbf{x}^s$ be an image from $\mathcal{D}_s$ and $\mathbf{x}^{\phi} = \phi(\mathbf{x}^s)$ be the transformation of $\mathbf{x}^s$, denoted as $\mathbf{x}^{\phi}$, where $\phi \sim \Phi$. The model $\mathcal{F}_{det}$ is domain invariant if:

\begin{equation} 
    %P(y^s_n|x^s) = P(y^{\phi}_n|x^{\phi}) 
    \hat{\mathbf{p}}_n^{s} = \hat{\mathbf{p}}_n^{\phi}
    \label{defeq1}
\end{equation} 
\begin{equation} 
    1 - \mathrm{IoU}(\hat{\mathbf{b}}^s_n, \hat{\mathbf{b}}^{\phi}_n) = 0 \label{defeq2}
\end{equation}
  \end{definition}

\subsection{Proposed Method}

Through developing realizations of Eq. (\ref{defeq1}) and Eq. (\ref{defeq2}), we intend to improve the domain generalization of object detectors, when the training data from only a single source domain is available. So, our method has two main components: 1) diversification of the source domain (sec.~\ref{subsubsection:Diversifying Single Source Domain}) and 2) a detection alignment mechanism for improving generalization and calibration of detectors (sec.~\ref{subsubsection:Aligning the Augmented Detections}). Fig.~\ref{fig:architecture_new1} displays the overall architecture of our proposed method.

\subsubsection{Diversifying Single Source Domain}\label{subsubsection:Diversifying Single Source Domain}
Our proposed approach is motivated by various techniques used in diversifying source domains for object classification \cite{shankar2018generalizing, volpi2018generalizing, zhou2020learning, khan2021mode}. Deep convolutional networks often learn shortcuts instead of actual semantics, but diversifying training domains can reduce this problem. Since obtaining multiple domains is difficult, we simulate them through data augmentation. This introduces diversity in the training data and improves the network's ability to learn robust object representations.

We achieve this by perturbing the single source domain images using a set of existing visual corruptions $\Phi$. During training, we augment every image in mini-batch by using image augmentation function $\phi(.)$ which randomly samples a corruption method from $\Phi$ and apply to the input image. We then pass both original clean images along with their augmented versions to the model.
Specifically, we investigate the visual corruptions set used in image classification domain \cite{Cugu2022acvc} that simply combines ImageNet-C with Fourier transform-based corruptions, thereby providing a total of 22 corruptions with 5 severity levels.

\begin{figure}[t]
\centering
\includegraphics[width=\linewidth]{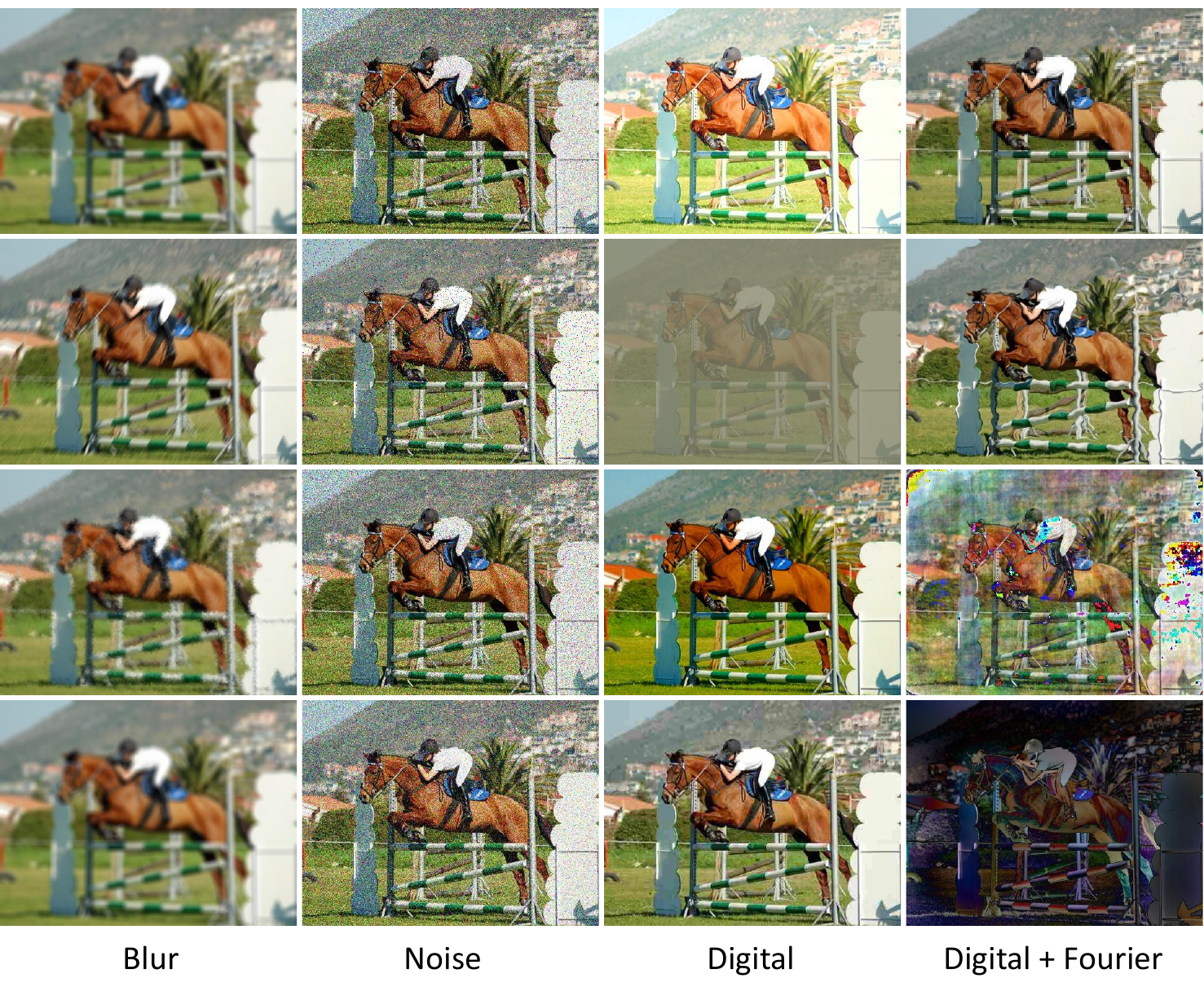}
\vspace{-0.5cm}
\caption{Examples of augmentations for domain diversification. 
\vspace{-0.6cm}}\label{fig:augs_samples}
\end{figure}

We classify image corruptions used for augmentation into five groups: Weather, Noise, Blur, Digital, and Fourier. In the case of Blur corruption, we employ various blur functions such as Gaussian, motion, glass, and defocus to smooth the pixel values in the image. Noise introduces different types of noise, including shot, impulse, speckle, and Gaussian, to perturb the image. Digital corruptions modify the pixel intensities by adjusting brightness, contrast, and saturation, or by changing the image resolution using JPEG compression, pixelation, and elastic transformation. Fig.~\ref{fig:augs_samples} shows the impact of such augmentations.
To ensure a fair comparison, we exclude weather-based transformations from our augmentation pool. Fourier-based corruptions such as phase scaling, constant amplitude, and High Pass Filter exploit the well-known property of Fourier transformation in that the phase component represents high-level semantics of an image, while the amplitude component covers low-level statistics or texture information. Perturbing these components enables us to reconstruct a new transformed image. Phase scaling introduces various artifacts by scaling the phase component in the Fourier spectrum. Constant amplitude replaces the amplitude component with a fixed vector, resulting in the loss of textual information and retaining only high-level semantics. We exclude the constant amplitude augmentation as it destroys the instance-level features. We note that this simple technique of diversifying source domain during object detector training outperforms baselines and competing methods by visible margins in challenging out-of-domain scenarios. As a result, though simple, as it makes minimal modifications to the baseline, it provides a solid baseline in single domain-generalized object detection.  

\subsubsection{Aligning Diversified Domains }\label{subsubsection:Aligning the Augmented Detections}
While the single-source domain diversification approach provides a strong baseline, it does not enforce the main requirement for the Domain invariance (Def.  \ref{def:invariant_def}); 
that is the predictions on diversified images $x^{\phi}$ do not align with predictions on original images $x^{s}$. To quantify this, we measured mAP of the model predictions over original images, and mAP of the same model's predictions on diversified images. Regardless of whichever diversification is performed, there is considerable performance gap, indicating the lack of generalization. This is depicted in Fig.~\ref{fig:augs_map}.
This gap can be explained when considering the following two constraints. 
(1.) \textit{Object Classification Constraint}, Eq.(\ref{defeq1}), states that model's predicted class probability distribution for the object in the source image, $x^{s}$, should be same for the corresponding object in diversified image $x^{\phi}$.
(2.) \textit{Object Localization Constraint}, Eq.(\ref{defeq2}), states that is the objects should be predicted at same location (and of same size) on both diversified and original version of the images. 
Note that here these constraints are not related to the accuracy of the prediction, but that any prediction made should be same. 

Unlike the image classification, where there is only one class prediction vector, enforcing such constraints in object detection is more challenging. We propose a strategy to overcome this challenge. 
Based on definition \ref{def:invariant_def}, on domain-invariance for object detection, we align the detections from two views i.e. clean and diversified (augmented) images both from the classification and the regression outputs. 
We take the widely used two-stage object detector as an example i.e Faster R-CNN. Given $\mathbf{x}_i^s$ or $\mathbf{x}_i^\phi$ as input, we can obtain a feature map $F \in \mathbb{R}^{m \times w \times h}$ from the backbone, where $m$, $h$, and $w$ denote the number of channels, height and width of the feature map, respectively. This feature map is then passed through region proposal network (RPN) to obtain object proposals $O \in \mathbb{R}^{Z \times 4}$, where $Z$ is the number of proposals. Next, RoI-Alignment (RA) is performed using $O$ i.e $A = \mathrm{RA}(O, F) \in \mathbb{R}^{Z\times m}$  to obtain a fixed length feature representation for each proposal. Detection loss is given by:

\begin{figure}[!t]
\centering
\includegraphics[width=\linewidth]{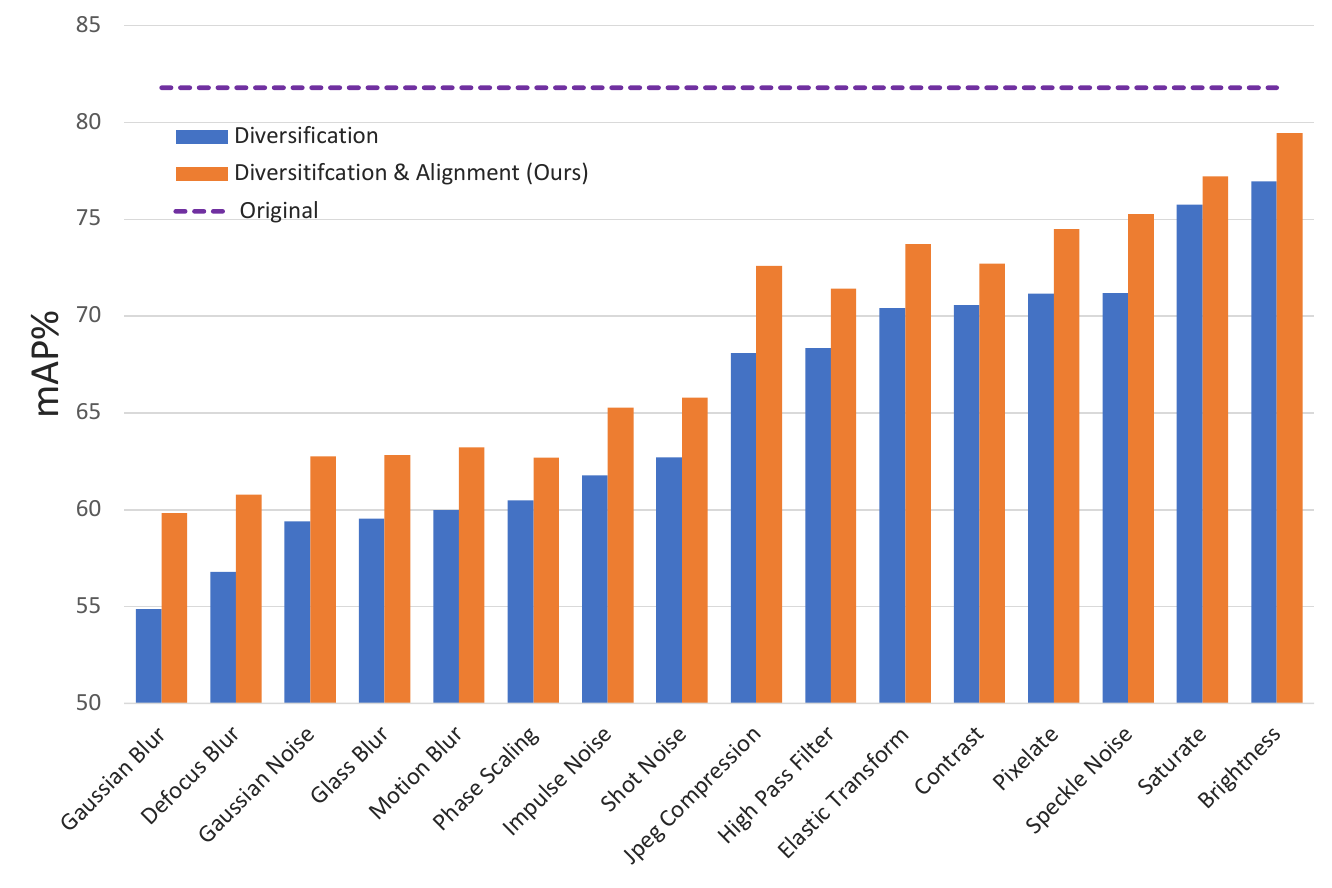}
% \vspace{-0.5cm}
\caption{We augment the validation set from the source domain by one augmentation at a time and report the performance of the strong baseline model trained on all these augmentations. There is noticeable gap between the performance on original and diversified images. Our alignment losses allows reducing the gap between the performance on original and diversified images. }
\vspace{-0.2cm}
\label{fig:augs_map}
\end{figure}
 
\begin{equation} \label{eq:3}
    \mathcal{L}_{\mathrm{det}} = \sum_{n=1}^{Z}L_{det} (\Psi(A_n), \mathbf{y}_n).
\end{equation}
Here $\Psi$ indicates the classifier and a bounding box regressor and $\mathbf{y}_n)$ is the ground truth including class label and bounding box coordinates for the $n^{th}$ proposal.
The $L_{det}$ consists of smooth-L1 loss and the negative log-likelihood loss.

\noindent\textbf{Aligning classification:}
To enforce \textit{Object Classification constraint for object detector generalization}, we propose to measure the Kullback-Leibler (KL) divergence between the classifier output probability vectors $\hat{\mathbf{p}}^s_n$ and $\hat{\mathbf{p}}_n^\phi$ corresponding to $\mathbf{x}_i^s$ and $\mathbf{x}_i^\phi$. % 

It is likely that RPN will produce a different set of proposals from $\mathbf{x}_i^s$ and $\mathbf{x}_i^\phi$ i.e. $O^s \neq O^\phi$. These proposals might not be spatially consistent. 
To achieve 1-1 correspondence between the proposals from $\mathbf{x}_i^s$ and $\mathbf{x}_i^\phi$, we perform predictions using the region proposals from the $\mathbf{x}_i^s$ and features pooled from the feature map for $\mathbf{x}_i^\phi$.
That is we pass the feature map $F^\phi$ corresponding to $\mathbf{x}_i^\phi$ with proposals from $O^s$: $A^{\phi'} = \mathrm{RA}(O^s, F^\phi)$. 
 $O^s$: $A^{\phi'} = \mathrm{RA}(O^s, F^\phi)$, obtaining $\hat{\mathbf{p}}_n^{\phi'}$.
The difference between the predictions is minimized using the KL-divergence: 

\begin{equation} \label{eq:5}
\mathcal{L}_{\mathrm{cal}} = \sum_{n=1}^{Z} \mathrm{KL}(\hat{\mathbf{p}}^s_n \| \hat{\mathbf{p}}_n^{\phi'}). 
\end{equation}

\noindent\textbf{Aligning localization:}
To make the \textit{Object Localization constraint for generalization} part of the training process,
we maximize the IoU between $\hat{\mathbf{b}}_n^s$ 
and 
$\hat{\mathbf{b}}_n^{\phi'}$. 
This is realized by minimizing the L2-squared norm between $\hat{\mathbf{b}}^s_n$ and $\hat{\mathbf{b}}_n^{\phi'}$:

\begin{equation} \label{eq:5}
\mathcal{L}_{\mathrm{ral}}= \| \hat{\mathbf{b}}_n^s - \hat{\mathbf{b}}_n^{\phi'} \|^2_2 
\end{equation}
Our overall training objective is given by: 
\begin{equation}
 % \begin{aligned}    
\mathcal{L}_{\mathrm{tot}} = \mathcal{L}_{\mathrm{det}} + \alpha  \mathcal{L}_{\mathrm{cal}} + \beta \mathcal{L}_{\mathrm{ral}}
%\end{aligned} 
\label{eq:total_loss}
\end{equation}
where $\alpha$ and $\beta$ are the hyperparameters for balancing the contributions of $\mathcal{L}_{\mathrm{cal}}$ and $\mathcal{L}_{\mathrm{ral}}$.

%% file: sec/4_experiments.tex
\section{Experiments}\label{sec:experiments}

\noindent\textbf{Datasets:}
  
\noindent\textbf{Real to Artistic} consists of four datasets, including the Pascal VOC which contains real images from 20 different object classes and acts as a source domain, and remaining three are Clipart1k, Watercolor2k and Comic2k which offer artistic and comic images and are used as target domains. 
Clipart1k shares the same 20 classes as Pascal VOC, while Watercolor2k and Comic2k consist of 6 classes each, which are subsets of the Pascal VOC classes. Following \cite{inoue2018cross}, we combine Pascal VOC2007 and VOC2012 trainval set resulting in 16551 images for training and 5K images from VOC2007 test for in-domain evaluation. The Clipart1k consists of 1K and Watercolor2k and Comic2k each contain 2K images which are equally split into training and test sets. Following \cite{inoue2018cross}, we select all images from Clipart1k and test sets from Watercolor2k and Comic2k for out-domain evaluations.
\noindent\textbf{Urban Scene Detection} is a self-driving dataset introduced by \cite{wu2022single}. It offers images of 5 different weather conditions including Daytime Sunny, Night Clear, Night Rainy, Dusk Rainy and Daytime Foggy .The images are collected from three different datasets: Barkeley Deep Drive 100k (BDD-100k)\cite{yu2018bdd100k}, FoggyCityscapes\cite{Cordts2016Cityscapes} and Adverse-Weather\cite{Hassaballah2021adverse}. Rainy scenes are synthetic images generated by simulating rain on BDD-100k images. Daytime Sunny serves as a source domain consisting 26,518 images out of which 8,313 are used for in-domain evaluation whereas Night Clear, Night Rainy, Dusk Rainy and Daytime Foggy are used as target domains containing 26,158, 2,494, 3,501, and 3,775 images, respectively. \\
\noindent\textbf{Implementation and Training Details:}
For two-stage object detector as a baseline, we use Faster R-CNN\cite{ren2015faster} with ResNet101\cite{he2016deep} as feature backbone, while for single-stage detector, we use FCOS \cite{tian2019fcos} with ResNet50-FPN\cite{DBLP:conf/cvpr/LinDGHHB17} as feature backbone. In all experiments our models are trained using SGD for 18k iterations with initial learning rate of 0.2 and dropped by a factor of 10 after 12K and 16k iterations. The batch size is set to 16. All results are reported using mean average precision (mAP) at IoU@0.5. For reporting calibration performance, we use the detection expected calibration error (D-ECE) \cite{kueppers_2020_CVPR_Workshops} and reliability diagrams \cite{guo2017calibration}.

\noindent\textbf{Real to artistic:}
We evaluate our method on challenging real to artistic domain-shift where the distribution shift is large i.e. from PASCAL VOC to Clipart1k, Watercolor2k, and Comic2k. Table \ref{tab_voc} shows that diversifying the training domain helps the model to generalize quite well by obtaining a significant gain of 8.5\%, 8.5\%, and 5.3\% from the Faster R-CNN baseline in Clipart1k, Watercolor2k, and Comic2k, respectively. With our proposed alignment losses, we are able to boost the overall performance by 13.2\%, 12.9\%, and 14.3\%. We also evaluated Normalization Perturbation (NP) \cite{fan2023towards} method on this benchmark and found it inferior to our method. Tables ~\ref{tab:cls_watercolor} and \ref{tab:cls_comic2k} reveal the class-wise results in Watercolor2k and Comic2k shifts. See supplementary for class-wise results on Clipart1k.

\begin{table}[t]
\centering
\scalebox{0.8}{
\begin{tabular}{|l|c|ccc|}
\hline
\rowcolor[HTML]{D6D6D6}
Method & VOC  & Clipart & Watercolor  & Comic\\
\hline
Faster R-CNN    & 81.8& 25.7& 44.5& 18.9  \\
NP \cite{fan2023towards}    & 79.2& 35.4& 53.3& 28.9  \\
Diversification (div.)    & 82.1& 34.2& 53.0& 24.2  \\

div. + $ \mathcal{L}_{\mathrm{cal}}$        & 82.1& 36.2& 53.9& 28.7  \\
div. + $ \mathcal{L}_{\mathrm{ral}}$        & 80.7& 35.0& 53.8& 28.7  \\

div. + $ \mathcal{L}_{\mathrm{cal}} + \mathcal{L}_{\mathrm{ral}}$ (Ours)      & 80.1& \textbf{38.9}& \textbf{57.4}& \textbf{33.2}  \\
\hline
\end{tabular}
} 
\caption{Performance comparison with baseline and possible ablations, mAP@0.5(\%) reported. The model is trained on Pascal VOC and tested on Clipart1k, Watercolor2k and Comic2k.}\label{tab_voc}
\end{table}

\begin{table}[t]

\scalebox{0.7}{
\begin{tabular}{|l|llllll|l|}
\hline
\rowcolor[HTML]{D6D6D6}
Method & bike& bird& car& cat& dog& person& mAP\\
\hline
Faster R-CNN & 85.7& 42.5& 36.4& 29.0& 18.7& 54.5& 44.5 \\
Diversification (div.) & 87.1& 51.7&\textbf{53.6}& 35.1& 23.6& 63.6& 52.5 \\
div. + $\mathcal{L}_{\mathrm{cal}}$       & 83.9& 50.4& 48.5& 40.3& \textbf{36.5}& 63.3& 53.8 \\

div. + $\mathcal{L}_{\mathrm{ral}}$       & 83.3& \textbf{52.3}& 50.7& 39.2& 32.7& 65.2& 53.9 \\

div. + $\mathcal{L}_{\mathrm{cal}} + \mathcal{L}_{\mathrm{ral}}$ (Ours) & \textbf{90.4}& 51.8& 51.9& \textbf{43.9}& 35.9& \textbf{70.2}& \textbf{57.4} \\
\hline
\end{tabular}
}

\caption{Class-wise AP(\%) comparison of baseline and proposed method on Pascal VOC to Watercolor2k scenario.}\label{tab:cls_watercolor}
\end{table}

\begin{table}[t]

\scalebox{0.7}{
\begin{tabular}{|l|llllll|l|}
\hline
\rowcolor[HTML]{D6D6D6}
Method& bike& bird& car& cat& dog& person& mAP\\
\hline
Faster R-CNN & 39.7& 9.1& 23.9& 9.1& 9.1& 22.2& 18.9 \\

Diversification (div.) & 41.7& 12.3& 29.0& 13.2& 20.6& 36.5& 25.5 \\

div. + $\mathcal{L}_{\mathrm{cal}}$       & 49.0& 15.8& 27.0& 20.2& 21.8& 38.4& 28.7 \\

div. + $\mathcal{L}_{\mathrm{ral}}$ & 53.3& 13.5& 29.0& 19.1& 20.5& 36.7&28.7 \\

div. + $\mathcal{L}_{\mathrm{cal}} + \mathcal{L}_{\mathrm{ral}}$ (Ours) & \textbf{54.1}& \textbf{16.9}& \textbf{30.1}& \textbf{25.0}& \textbf{27.4}& \textbf{45.9}& \textbf{33.2} \\
\hline
\end{tabular}
}

\caption{Class-wise AP(\%) comparison of baseline and proposed method on Pascal VOC to Comic2k scenario.}\label{tab:cls_comic2k}
\end{table}

\noindent\textbf{Urban scene detection:}
Table \ref{tab_multiweather} shows the results on different weather conditions manifesting various shifts scenarios. Due to large shifts like Daytime-Foggy and Night Raining the objects become unclear. We note that, by only diversifying single domain, we obtain 0.5\%, 7\%, 6.3\%, and 2.5\% on Night Clear (NC), Dusk Rainy (DR), Night Rainy (NR), and Daytime Foggy (DF), respectively. Our method beats all baselines and state-of-the-art method from \cite{Vidit_2023_CVPR}. Over the baseline, our method delivers the highest gain of 8.4\% on Night Rainy while 8.1\%, 4.1\% and 3.6\% gains on Dusk Rainy, Daytime Foggy and Night Clear respectively. \\

\begin{table}[!htp]

\centering
\scalebox{0.82}{

\begin{tabular}{|l|l|llll|}
\hline
\rowcolor[HTML]{D6D6D6}
Method & DS   & NC  & DR   & NR & DF\\
\hline
Faster R-CNN    & 51.8& 38.9& 30.0& 15.7& 33.1  \\
SW\cite{pan2018switchable}              & 50.6& 33.4&26.3& 13.7& 30.8 \\
IBN-Net\cite{pan2018IBN}         & 49.7& 32.1&26.1& 14.3& 29.6 \\
IterNorm\cite{Huang2019iternorm}        & 43.9& 29.6&22.8& 12.6& 28.4 \\
ISW\cite{choi2021ISW}             & 51.3& 33.2&25.9& 14.1& 31.8 \\

Wu \etal \cite{wu2022single}        & \textbf{56.1}& 36.6&28.2& 16.6& 33.5 \\

Vidit \etal \cite{Vidit_2023_CVPR}  & 51.3& 36.9& 32.3& 18.7& \textbf{38.5} \\
\hline

Diversification    & 50.6& 39.4& 37.0& 22.0& 35.6  \\
Ours     & 52.8& \textbf{42.5}& \textbf{38.1}& \textbf{24.1}& 37.2  \\
\hline
\end{tabular}
 }

\caption{Results, mAP@0.5(\%) reported on a multi-weather scenario where the model is trained on Daytime Sunny (DS) and tested on Night-Clear (NC), Night-Rainy (NR), Dusk-Rainy (DR) and Daytime-Foggy (DF). The results for SW, IBN-Net, IterNorm, ISW, S-DGOD are directly taken from \cite{wu2022single}.}\label{tab_multiweather}
\end{table}

\begin{figure*}[!htp]
\centering
\begin{subfigure}[t]{0.47\linewidth}
    \includegraphics[width=\linewidth]{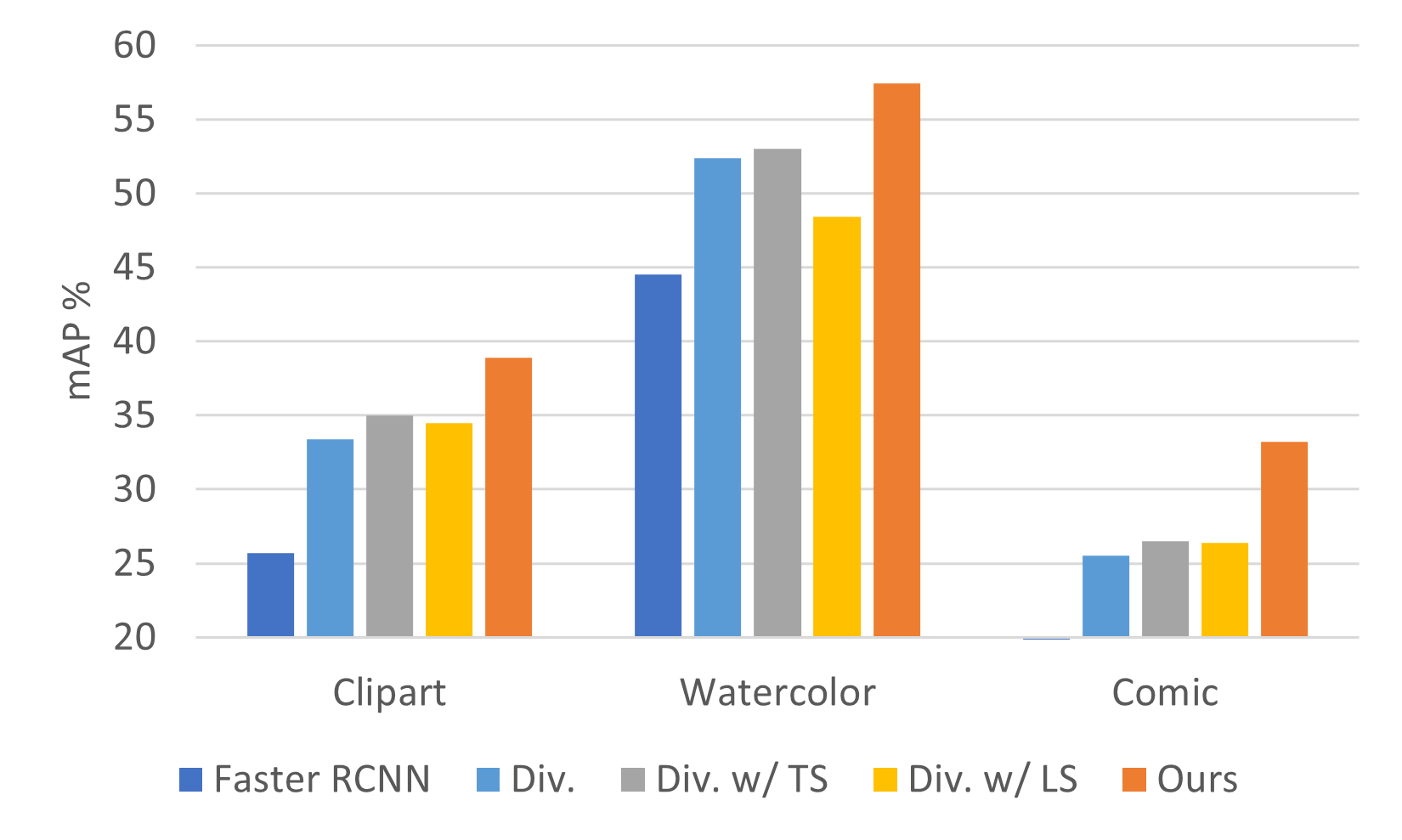}
  \end{subfigure}%
  \hspace{\fill}  
  \vspace{\fill}
  \begin{subfigure}[t]{0.47\linewidth}
  
   \includegraphics[width=\linewidth]{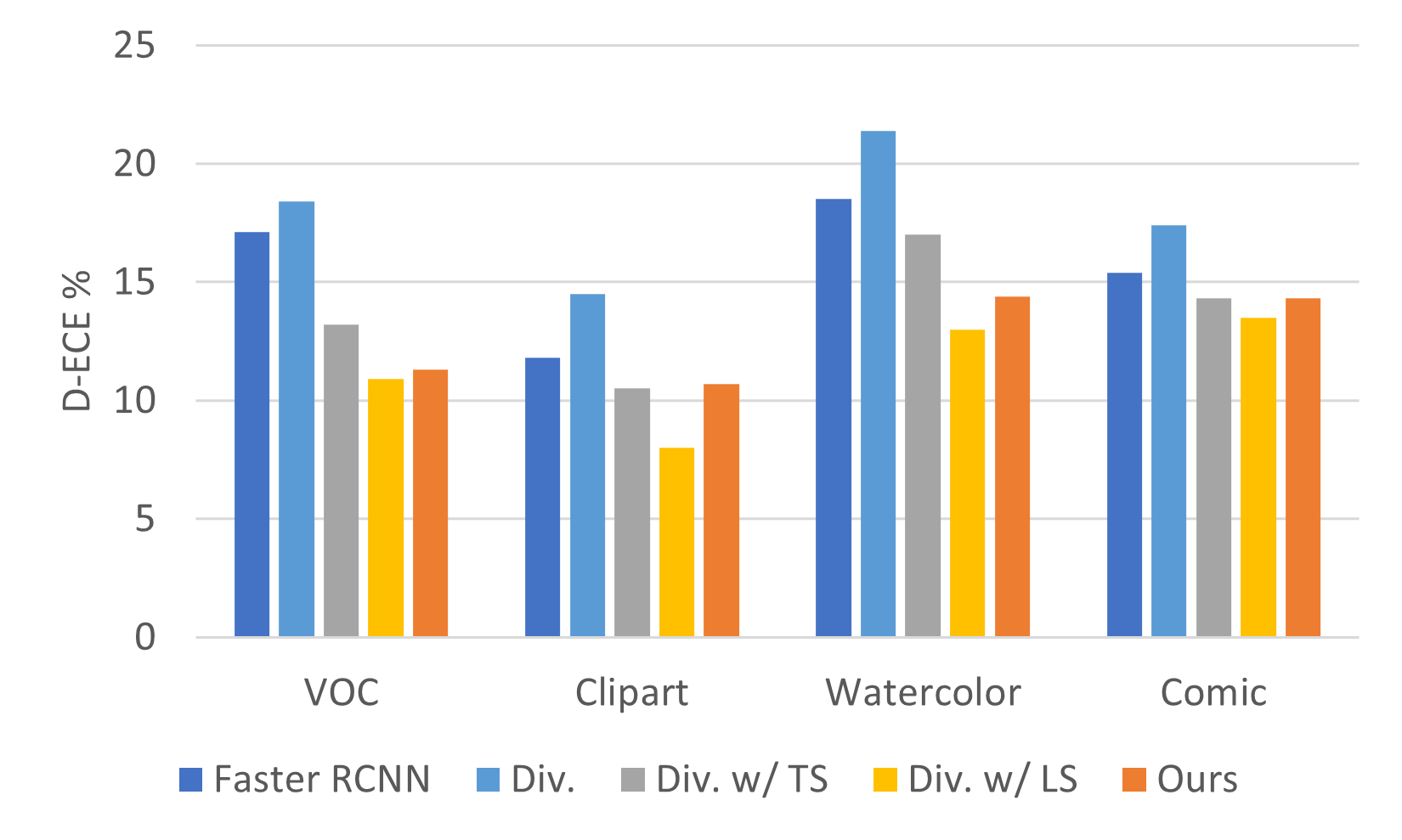}

  \end{subfigure}
\caption{Our method (diversification and alignment) results in both considerable improvement in  domain generalization and out-of-domain calibration. Diversification with the Label Smoothing (LS) or Temperature Scaling (TS) improves calibration but overall lower mAP indicates lacking in generalization. Note that our method does not have an explicit model calibration mechanism. (left) mAP: higher the better, (right) D-ECE: lower the better. 
\label{fig:voc_map_ece}
 }
 \vspace{-0.3cm}
\end{figure*}

\begin{figure}[h]
\centering
\includegraphics[width=\linewidth]{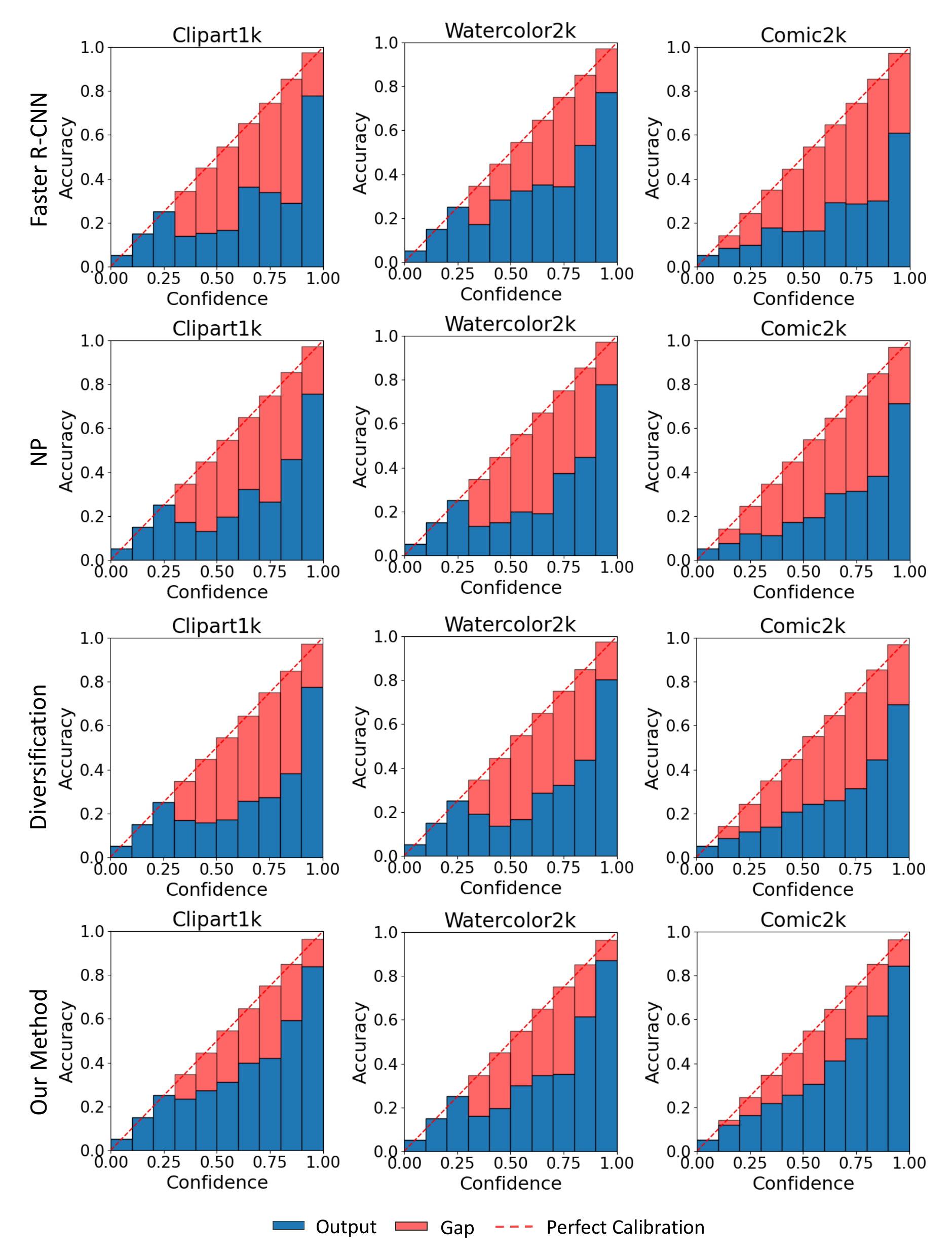}

%\vspace{-0.2cm}
\caption{Reliability Diagram for different target domains.}%, our method considerably improves the model calibration.}
\label{fig:reliability_diagrams}
\vspace{-1em}

\end{figure}

\noindent \textbf{Model Calibration:}

Table \ref{tab_ece_voc},
reports the calibration performance (D-ECE) of the base model and the impact of using our diversification and alignment approach under various shift scenarios. As seen, simple diversification of the domain, while improving generalization, negatively affects the calibration for both in-domain and out-domain scenarios. 
This is to be expected since model is trained using Negative log-likelihood loss in $\mathcal{L}_{\mathrm{det}}$ (Eq.~\ref{eq:3}) which is in general attributed to making models highly confident, resulting in poor calibration. On the diversified images, where information might have been reduced or content has become noisy (w.r.t original image), high confidence predictions affects the calibration. In contrast, as detailed in (Sec. \ref{subsubsection:Aligning the Augmented Detections}), our alignment step implicitly reduces the overconfidence in predictions (Fig.~\ref{fig:reliability_diagrams}). Overall, as our results show, we achieve not only better generalization (higher mAP) but also better calibration (lower D-ECE).
\begin{table}[!htp]
\centering
\scalebox{0.68}{
\begin{tabular}{|l|ccc|cccc|}
\hline
& \multicolumn{3}{c|}{Artistic Shifts}                     & \multicolumn{4}{c|}{Urban Scene}                                                 \\ \hline
\rowcolor[HTML]{D6D6D6}Method & \multicolumn{1}{c}{Clipart}                        & \multicolumn{1}{c}{Watercolor}                     & Comic                          & \multicolumn{1}{l}{NR} & \multicolumn{1}{l}{DR} & \multicolumn{1}{l}{NC} & DF \\ \hline
Faster R-CNN                                       & \multicolumn{1}{c}{11.9}                           & \multicolumn{1}{c}{18.5}                           & 15.4                            & 31.5& 29.3& 27.9& 25.8   \\ 
Diversification (div.)                             & \multicolumn{1}{c}{14.5}                           & \multicolumn{1}{c}{21.4}                           & 17.4                           & 33.0& 30.2& 28.9& 25.7    \\ 
Ours                                               & \multicolumn{1}{c}{\textbf{10.7}} & \multicolumn{1}{c}{\textbf{14.4}} & \textbf{14.3} & \textbf{29.3}& \textbf{24.9}& \textbf{15.8}& \textbf{20.6}    \\  \hline
\end{tabular}
   }
   
\caption{Comparison of calibration performance using D-ECE metric (\%) on Real to artistic shifts and in urban scene detection. }\label{tab_ece_voc} 
\vspace{-0.2cm}
\end{table}
To further make apparent the the utility of our alignment approach in achieving better calibration, We also include the impact of using existing simple calibration and generalization techniques such as temperature scaling (TS) \cite{guo2017calibration} or label smoothing (LS) \cite{muller2019does} with our diversification approach. Fig.~\ref{fig:voc_map_ece} depicts the improvements our presented methods achieve in multiple shift scenarios.

\begin{figure*}[!t]
\centering
\includegraphics[scale=0.142]
{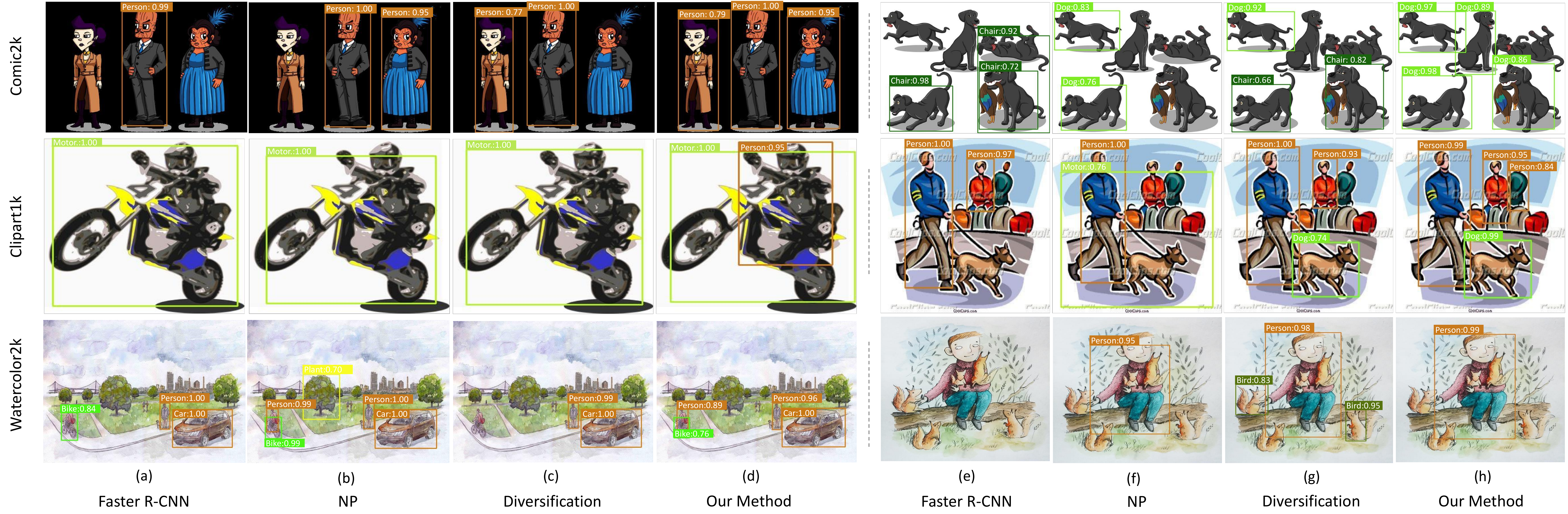}
 
\caption{Qualitative results of baseline (Faster-RCNN), only diversifying domain, and our method.}\label{fig:voc_qualitative} \vspace{-0.2em}
\end{figure*}

\noindent\textbf{Results with single-stage detector:}
To show the applicability of our proposed method, we instantiate it in one-stage detector, namely FCOS \cite{tian2019fcos} (Table \ref{tab_fcos}). In comparison to FCOS, our method delivers a significant gain of 13.0\%, 10.7\% and 15.8\% on Clipart1k, Watercolor2k, and Comic2k shifts, respectively.

\begin{table}[h]
\centering
\scalebox{0.8}{
\begin{tabular}{|l|c|ccc|}
\hline
\rowcolor[HTML]{D6D6D6}
Method & VOC  & Clipart & Watercolor  & Comic\\
\hline
FCOS                & 78.1& 24.4& 44.3& 15.4  \\
Diversification (div.)       & 79.6& 31.7& 48.8& 25.2  \\
div. + $ \mathcal{L}_{\mathrm{cal}}$         & \textbf{80.1}& 35.4& 52.6& 29.4  \\
div. + $ \mathcal{L}_{\mathrm{ral}}$         & 77.5& 29.8& 50.3& 24.0  \\
div. + $ \mathcal{L}_{\mathrm{cal}} + \mathcal{L}_{\mathrm{ral}}$ (Ours)           & 77.5& \textbf{37.4}& \textbf{55.0}& \textbf{31.2}  \\
\hline
\end{tabular}}

\caption{Performance comparison with single-stage baseline, mAP@0.5(\%) reported. The model is trained on Pascal VOC and tested on Clipart1k, Watercolor2k and Comic2k. }\label{tab_fcos}
\vspace{-0.2cm}
\end{table}

\noindent\textbf{Comparison with domain-adaptive detection methods:}
Table \ref{tab_da} compares our method with some unsupervised domain-adaptive object detectors using the same backbone. Note that these methods assume access to target domain and directly train using unlabelled target samples. In contrast even though our method does not have access to the target domain dataset at training time, it still shows better generalization results than many domain-adaptive detection methods. For instance, in watercolor and comic shifts, our model outperforms all methods by delivering a gain of 3.2\% and 2.5\%, respectively. Results for the existing domain adaptive methods are taken from \cite{chen2021dual}.  \\

\begin{table}[!htp]
\centering
\scalebox{0.8}{

\begin{tabular}{|l|ccc|}
\hline
\rowcolor[HTML]{D6D6D6}
Method & Clipart & Watercolor  & Comic\\
\hline
DA-Faster\cite{chen2018domain}   &19.8& 46.0& -  \\
SWDA\cite{saito2019strong}        &38.1& 53.3& 27.4  \\
HTCN\cite{chen2020harmonizing}        &40.3& - &  - \\
DBGL\cite{chen2021dual}        &\textbf{41.6}& 53.8& 29.7 \\
\hline
Our Method         &38.5& \textbf{57.0}& \textbf{32.2}  \\
\hline
\end{tabular}
}

\caption{\small Comparison with domain adaptive detectors in mAP@0.5(\%).} 
% \vspace{-0.1cm}
%on Pascal VOC to Clipart1k, Pascal VOC to Watercolor2k and Pascal VOC to Comic1k}
\label{tab_da}
% \end{SCtable}
\end{table}

\noindent\textbf{Qualitative Results:} We sample some results for the qualitative comparison in Fig.~\ref{fig:voc_qualitative}. Aligning the predicted confidence and predicted localization between the original and diversified image, results in improved generalizabilty as visible from columns (c \& f) where objects missed by baseline, diversification and NP\cite{fan2023towards} method, were properly localized and classified.  Note that diversification actually results in more false positive detections with high confidence than baseline. This is visible in the column g where objects have been falsely classified as chairs and birds.

\noindent\textbf{Effect of different types of augmentations:}
We analyze the impact of different types of visual corruptions obtained from $\Phi$ in our method. First, we only diversify VOC-dataset using ImageNet-C corruption and then with Fourier-based corruption.
As reported in Table~\ref{tab_augs}, ImageNet-C corruption has a slight edge over Fourier corruption,  however, using both provides the best results.  

This suggests that obtaining diversity through both types of corruption is complimentary. 

\begin{table}[t]
\centering
\scalebox{0.8}{
\begin{tabular}{|l|ccc|}
\hline
\rowcolor[HTML]{D6D6D6}
Method   & Clipart & Watercolor  & Comic\\
\hline
Faster R-CNN    & 25.7& 44.5& 18.9  \\
Ours - w/Fourier-based      & 36.1& 54.8& 29.1  \\
Ours - w/Imagenet-C         & 36.8& 55.5& 30.6  \\

Ours         & \textbf{38.9}& \textbf{57.4}& \textbf{33.2}  \\
\hline
\end{tabular}}

\caption{Ablation study with different augmentations used for diversifying source domain in our method, mAP@0.5(\%) reported.} 
\label{tab_augs}
\vspace{-0.1em}
\end{table}
\noindent\textbf{Aligning Classification vs aligning localization:} Not aligning one of the prediction tasks across the diversified domains results in considerable degradation of the results (Table~\ref{tab_voc}). 
In general, the results indicate that, as expected, the alignment of both these tasks is equally vital for achieving cross-domain generalizable object detectors.

%% file: sec/5_conslusion.tex
\section{Conclusion}\label{sec:Conclusion and Future Work}

We proposed a detector-agnostic, simple, and effective approach to improve the generalization and calibration of the object detectors given only a single source domain for training. To this end, we first explored effective image augmentation methods for domain diversification and developed a strong baseline for single-DGOD. Our results shows that by carefully selecting a set of augmentations, a base detector can outperform SOTA methods for single domain generalization by a good margin. 
Secondly, we proposed a novel method to align the detections across the original and diversified images. This alignment procedure leads to better generalized and well-calibrated object detectors, which are crucial for accurate decision-making in safety-critical applications. Extensive results on various challenging shift scenarios corroborate the effectiveness and applicability of our approach.

%% file: sec/X_suppl.tex
% \clearpage
\setcounter{page}{1}
\maketitlesupplementary

%%%%%%%%% BODY TEXT
\section{Additional Results}
\noindent\textbf{Sensitivity to hyperparameters:} We analyze the sensitivity of $\alpha$ and $\beta$ (from Eq.(6) in the main paper) in Tables \ref{tab_alpha} and \ref{tab_beta}, respectively. We observe that overall the performance of our method is robust to the choice of hyperparameters. It outperforms the baseline on all different settings of hyperparameters.

\begin{SCtable}[\sidecaptionrelwidth][h]
\scalebox{0.8}{
\begin{tabular}{|l|ccc|}
\hline
\rowcolor[HTML]{D6D6D6}
$\alpha$ & Clipart & Watercolor  & Comic\\
\hline
0.0   & 35.0            & 53.8          & 28.7          \\
0.1   & 36.1          & 53.6          & 27.1          \\
0.3   & 37.9          & 55.1          & 30.2          \\
0.5   & 38.0            & 55.7          & 29.7          \\
0.7   & 37.1          & 55.6          & 31.5          \\
1.0     & \textbf{38.9} & \textbf{57.4} & \textbf{33.2} \\
1.2     & 38.6 & 56.8 & 33.0 \\
\hline
\end{tabular}
\caption{Sensitivity analysis of hyperparameter $\alpha$ after setting $\beta$ to a fixed value of 1.0.}\label{tab_alpha}}
\end{SCtable}

\begin{SCtable}[\sidecaptionrelwidth][h]
\scalebox{0.8}{
\begin{tabular}{|l|ccc|}
\hline
\rowcolor[HTML]{D6D6D6}
$\beta$   & Clipart & Watercolor  & Comic\\
\hline
0.0    & 36.2          & 53.9          & 28.7          \\
0.1  & 36.1          & 54.8          & 30.2          \\
0.3  & 37.8          & 55.7          & 30.8          \\
0.5  & \textbf{39.4}          & 54.7          & 32.0            \\
0.7  & 37.3          & 56.8          & 32.2          \\
1.0  & 38.9 & \textbf{57.4} & \textbf{33.2} \\
1.2  & 37.7          & 56.3          & 31.7          \\
\hline
\end{tabular}
}
\caption{Sensitivity analysis of hyperparameter $\beta$ after setting $\alpha$ to 1.0.}\label{tab_beta}
\end{SCtable}

\noindent\textbf{Results with another diversification approach:}
A recent work dubbed as Normalization Perturbation (NP) \cite{fan2023towards}, which targets single-DGOD, proposed to diversify the single domain by perturbing the low-level channel statistics. We evaluated this method on Real to Artistic benchmark and report the result in Table~\ref{tab_np}. To further investigate the efficacy of our alignment losses, namely $\mathcal{L}_{\mathrm{cal}} + \mathcal{L}_{\mathrm{ral}}$ from sec 3.2.2 in the main paper, we replace our diversification technique (sec 3.2.1 main paper) with the one proposed in \cite{fan2023towards}. 
Table \ref{tab_np} shows that our alignment losses are effective even with a different diversification approach, however, the results are still inferior to our proposed method.

\begin{table}[!htp]
\centering
\scalebox{0.8}{

\begin{tabular}{|l|c|ccc|}
\hline
\rowcolor[HTML]{D6D6D6}
Method & VOC  & Clipart & Watercolor  & Comic\\
\hline
NP    & 79.2& 35.4& 53.3& 28.9  \\

NP + $  \mathcal{L}_{\mathrm{cal}} + \mathcal{L}_{\mathrm{ral}}$ & 77.9& 37.3& 56.0& 32.2  \\

 Ours      & \textbf{80.1}& \textbf{38.9}& \textbf{57.4}& \textbf{33.2}  \\
\hline
\end{tabular}
} 

\caption{ 
Results after replacing our diversification technique (sec 3.2.1 main paper) with the one proposed in NP \cite{fan2023towards}. The model is trained on Pascal VOC and tested on Clipart1k, Watercolor2k and Comic2k.}\label{tab_np}

\end{table}

\noindent\textbf{Results on Medical Dataset:}
We also evaluate our method under the domain shift in a medical imaging scenario (see Tables \ref{tab_LCM},\ref{tab_ece_lcm}). To this end, we show results on recently proposed Malaria detection M5 benchmark\cite{Sultani_2022_CVPR}.~This benchmark includes data from two domains i.e. images of blood-smear from different malaria patients taken by a high cost microscope (HCM) and a low cost microscope i.e (LCM). The HCM serves as the source domain and LCM is used as the target domain. Table \ref{tab_LCM} shows that our proposed method is capable of generalizing to an unseen medical imaging domain (LCM), and it outperforms several other domain-adaptive detectors. Also, Table \ref{tab_ece_lcm} reveals that, compared to baseline (Faster R-CNN) and only diversification, our method improves the calibration of out-of-domain detections. 

\begin{table}[h]
\centering
 \scalebox{0.89}{

\begin{tabular}{|l|c|c|}
\hline
\rowcolor[HTML]{D6D6D6}
Method & HCM & LCM \\
\hline
Xu et al.\cite{xu2020cross} & - & 15.5\\
Saito et al.\cite{saito2019strong} & - & 24.8\\
Chen et al.\cite{chen2018domain} & - & 17.6\\
Sultani et al.\cite{Sultani_2022_CVPR} & 66.8 & \textbf{37.5}\\
\hline
Faster R-CNN  &71.4& 15.1  \\
Diversification  &\textbf{74.7}& 25.0  \\
Our Method         &70.3& \textbf{35.9}  \\
\hline
\end{tabular}
 }

\caption{Performance comparison (mAP \%) under domain shift in medical imaging modality. Model is trained on HCM and tested on LCM (out-domain).} 
\label{tab_LCM}
\label{tab_da}
\end{table}

\begin{table}[!htp]
\centering
\scalebox{0.8}{
\begin{tabular}{|l|c|}
\hline
\rowcolor[HTML]{D6D6D6}
Method & LCM\\
\hline
Faster R-CNN    & 8.4  \\
Diversification    & 8.0 \\
Ours     & \textbf{5.5} \\
\hline
\end{tabular}
   }
\caption{Calibration performance using D-ECE metric (\%) under domain-shift in medical imaging modality. }\label{tab_ece_lcm} %\vspace{-1.5em}
\end{table}

\noindent\textbf{Additional results with FCOS:}
We also report results of our method with FCOS baseline on Urban scene detection benchmark (see Table~\ref{tab_multiweather_fcos}). We note that, our method is also effective with FCOS baseline on the challenging multi-weather domain shifts.

\begin{table}[!htp]

\centering
\scalebox{0.82}{

\begin{tabular}{|l|l|llll|}
\hline
\rowcolor[HTML]{D6D6D6}
Method & DS   & NC  & DR   & NR & DF\\
\hline

FCOS    & 40.8& 29.7& 22.1& 12.6& 25.3  \\
Diversification    & 48.3& 35.6& 32.6& 18.7& 31.3  \\
Ours     & \textbf{53.7}& \textbf{37.9}& \textbf{37.4}& \textbf{21.2}& \textbf{34.9}  \\
\hline
\end{tabular}
 }
 
\caption{Results (\%) of our method with FCOS baseline on multi-weather scenario where model is trained on Daytime Sunny (DS) and tested on Night-Clear (NC), Night-Rainy (NR), Dusk-Rainy (DR) and Daytime-Foggy (DF).}\label{tab_multiweather_fcos} 
\end{table}

\begin{figure*}[!htp]
\centering
\includegraphics[width=\linewidth]{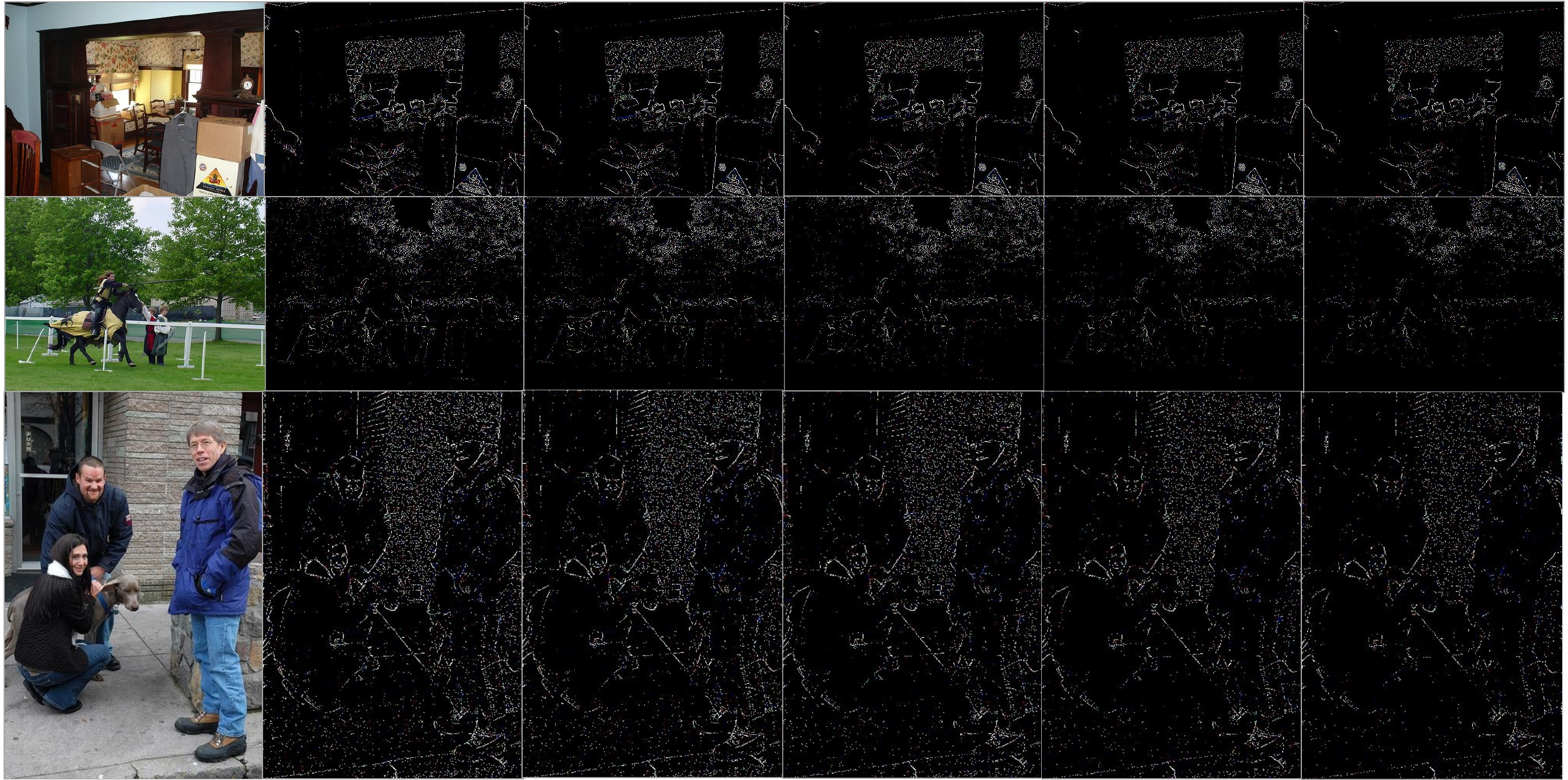}
\caption{Diversification results after applying constant amplitude visual corruption. We see that the objects information is mostly destroyed. Here the columns show the increasing severity levels.}

\label{fig:constant_amplitude}
\end{figure*}

\noindent\textbf{Diversification with constant amplitude:}
We omit the constant amplitude augmentation method from our final pool of visual corruptions for achieving diversification as it destroys the instance-level information which is crucial for object detection. Fig.~\ref{fig:constant_amplitude} shows some sample images when constant amplitude is applied. We see that the objects information is mostly destroyed. 

\begin{figure*}[!htp]
\centering
\includegraphics[width=\linewidth]{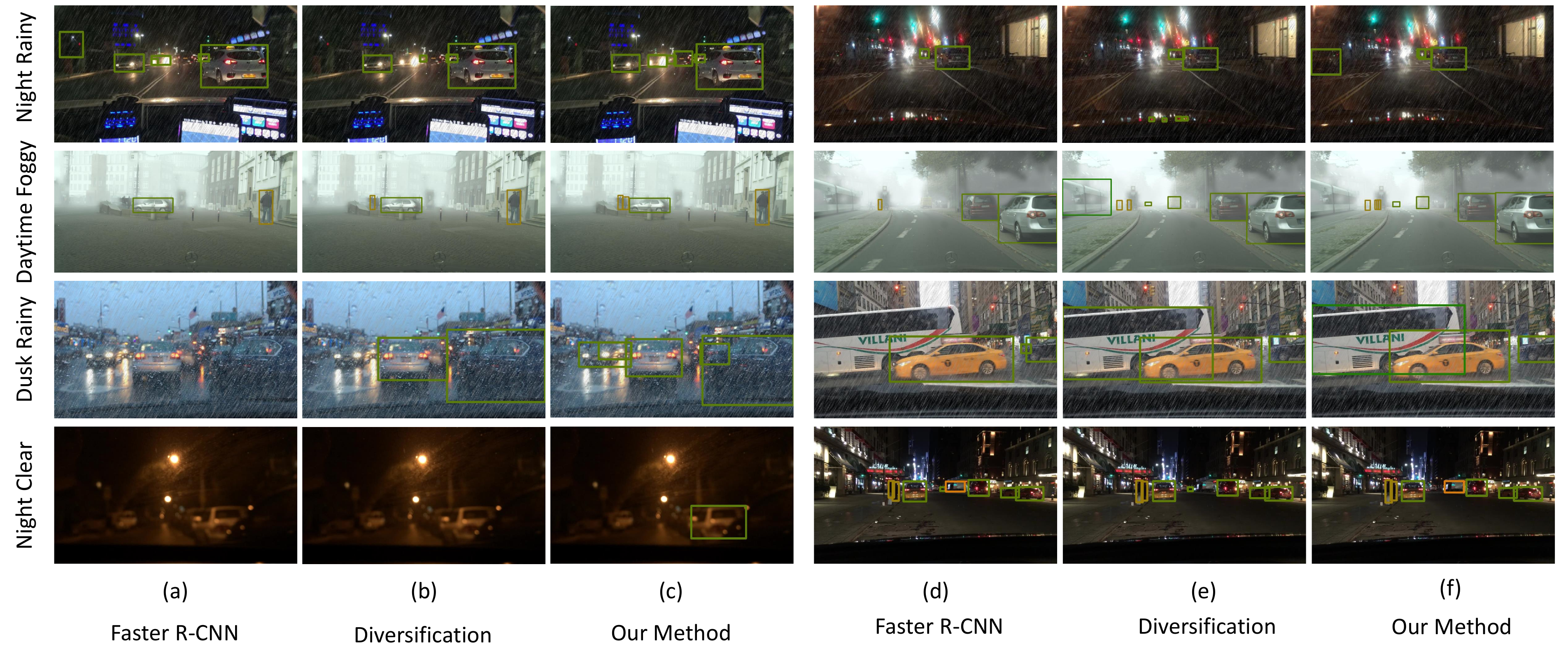}
\caption{Sampled results from urban scene detection benchmark. Our method accurately detects objects e.g \textcolor[HTML]{5b7f04}{car}, \textcolor[HTML]{D97C07}{truck}, \textcolor[HTML]{997e0a}{person}, \textcolor[HTML]{1F7F01}{bus}. For better viewing, higher zoom is recommended.}
\label{fig:qualitative_urban_scene}
\end{figure*}

\noindent\textbf{Additional qualitative results:}
We sample some results from urban scene detection benchmark in Fig.~\ref{fig:qualitative_urban_scene}. Our alignment losses allows detecting the object that were missed by baselines i.e Faster R-CNN and our diversification techniques as shown in columns (c, f).
Diversification result in more false positive detection as shown in column (e). For instance, in Night Rainy, Daytime Foggy and Night Clear, the background is detected as car and bus, and in dusk rainy, a bus is misclassfied as car.

\noindent\textbf{Class-wise performance on Clipart1k:}
Table \ref{tab:cls_clipart} shows the class-wise results when trained on Pascal VOC and evaluated on Clipart1k.

\begin{table*}[h]
\centering
\scalebox{.65}{
    
    \begin{tabular}{|l|llllllllllllllllllll|l|}
    \hline
\rowcolor[HTML]{D6D6D6}
    
        Method & place & bike & bird & boat & bottle & bus & car & cat & chair & cow & table & dog & horse & motor. & person & plant & sheep & sofa & train & tv & mAP \\ \hline
        Faster R-CNN & 19.9 & 51.6 & 17.0 & 21.9 & 27.2 & 49.6 & 25.5 & 9.1 & 35.1 & 9.1 & 25.4 & 3.0 & 29.2 & 48.9 & 30.1 & 40.3 & 9.1 & 6.7 & 35.2 & 21.0 & 25.7 \\
        Diversification (div.) & 29.3 & 50.9 & 23.4 & \textbf{35.3} & 45.3 & 49.8 & \textbf{33.4} & \textbf{10.6} & 43.3 & 22.3 & 31.6 & 4.5 & 32.9 & 51.9 & 40.2 & 51.1 & 18.2 & 29.6 & 42.3 & 28.5 & 33.7 \\ 
        div. + $\mathcal{L}_{\mathrm{cal}}$  & 30.4 & 61.7 & 23.8 & 28.4 & 41.6 & 51.7 & 32.6 & 7.6 & \textbf{46.9} & 31.4 & \textbf{36.8} & 8.3 & 41.7 & 57.2 & 48.4 & 48.5 & 18.2 & 38.2 & 40.1 & 30.4 & 36.2 \\
        
        div. + $\mathcal{L}_{\mathrm{ral}}$  & 34.0 & 56.2 & \textbf{24.4} & 28.9 & 41.2 & 50.4 & 32.6 & 4.5 & 45.9 & 29.5 & 30.2 & 12.8 & 39.4 & \textbf{59.5} & 46.4 & 50.1 & \textbf{18.2} & 26.2 & \textbf{43.4} & 26.2 & 35.0 \\ 
        div. + $\mathcal{L}_{\mathrm{cal}} + \mathcal{L}_{\mathrm{ral}}$ (Ours) & \textbf{34.4} & \textbf{64.4} & 22.7 & 27.0 & \textbf{45.6} & \textbf{59.2} & 32.9 & 7.0 & 46.8 & \textbf{55.8} & 28.9 & \textbf{14.5} & \textbf{44.4} & 58.0 & \textbf{55.2} & \textbf{52.1} & 14.8 & \textbf{38.4} & 42.5 & \textbf{33.9} & \textbf{38.9} \\ \hline
    \end{tabular}
}

\caption{Class-wise AP(\%) comparison of baseline and proposed method on Pascal VOC to Clipart1k scenario.}\label{tab:cls_clipart}
\end{table*}

